\pdfoutput=1
\documentclass[sigconf]{acmart}

\usepackage{graphicx}
\usepackage{booktabs}
\usepackage{amsmath,bm}
\usepackage{multirow}
\usepackage{tabularx}
\usepackage{xcolor,colortbl}
\usepackage{subcaption}
\usepackage{listings}
\usepackage{tcolorbox}
\usepackage{cleveref}
\usepackage{pifont}
\usepackage{relsize}

\usepackage{xurl}

\title{PTEI: Integrating Personality Traits to Enhance Emotional Intelligence in Large Language Models}

\author{Amir Reza Jafari}
\orcid{0009-0009-6297-1885}
\affiliation{%
	\institution{Samovar, Telecom SudParis, Institut Polytechnique de Paris}
	\city{Palaiseau}
	\country{France}
}
\author{Praboda Rajapaksha}
\authornotemark[2]
\affiliation{%
	\institution{Department of Computer Science, Aberystwyth University}
	\city{Aberystwyth}
	\country{Wales}
}
\author{Reza Farahbakhsh}
\affiliation{%
	\institution{Samovar, Telecom SudParis, Institut Polytechnique de Paris}
	\city{Palaiseau}
	\country{France}
}
\author{Noel Crespi}
\affiliation{%
	\institution{Samovar, Telecom SudParis, Institut Polytechnique de Paris}
	\city{Palaiseau}
	\country{France}
}


\begin{document}

\begin{abstract}
Despite advances in \textit{Emotional Intelligence (EI)}, Large Language Models (LLMs) still significantly underperform humans in complex emotional reasoning. This gap originates partly from the limited incorporation of individual differences, particularly personality traits, which are fundamental to human emotional inference. To address this, we propose \textbf{PTEI}, a novel framework for integrating Personality Traits into Emotional Intelligence tasks using LLMs. In PTEI, MBTI and OCEAN personality traits are first extracted directly from the given emotional scenarios and then utilized as contextual knowledge within personality-aware prompts, guiding LLMs to accurately infer emotions and their underlying causes. To ensure optimal contextual grounding, we employ \textit{Contrastive Learning} to construct an optimized retrieval system that surfaces emotionally and personally aligned scenarios, enhancing reasoning quality. Extensive experiments on established EI benchmarks show that PTEI enhances Emotional Understanding (EU) capabilities of various LLMs in EI, with the strongest improvement observed in GPT models, where combining PTEI with \textit{Chain-of-Thought (CoT) reasoning} yields an additional 4\% increase in accuracy. These findings underscore PTEI's contribution toward advancing AI systems with more sophisticated social and psychological grounding.
\end{abstract}

\maketitle

\section{Introduction}

\begin{figure}[ht]
    \centering
    \includegraphics[width=0.5\textwidth]{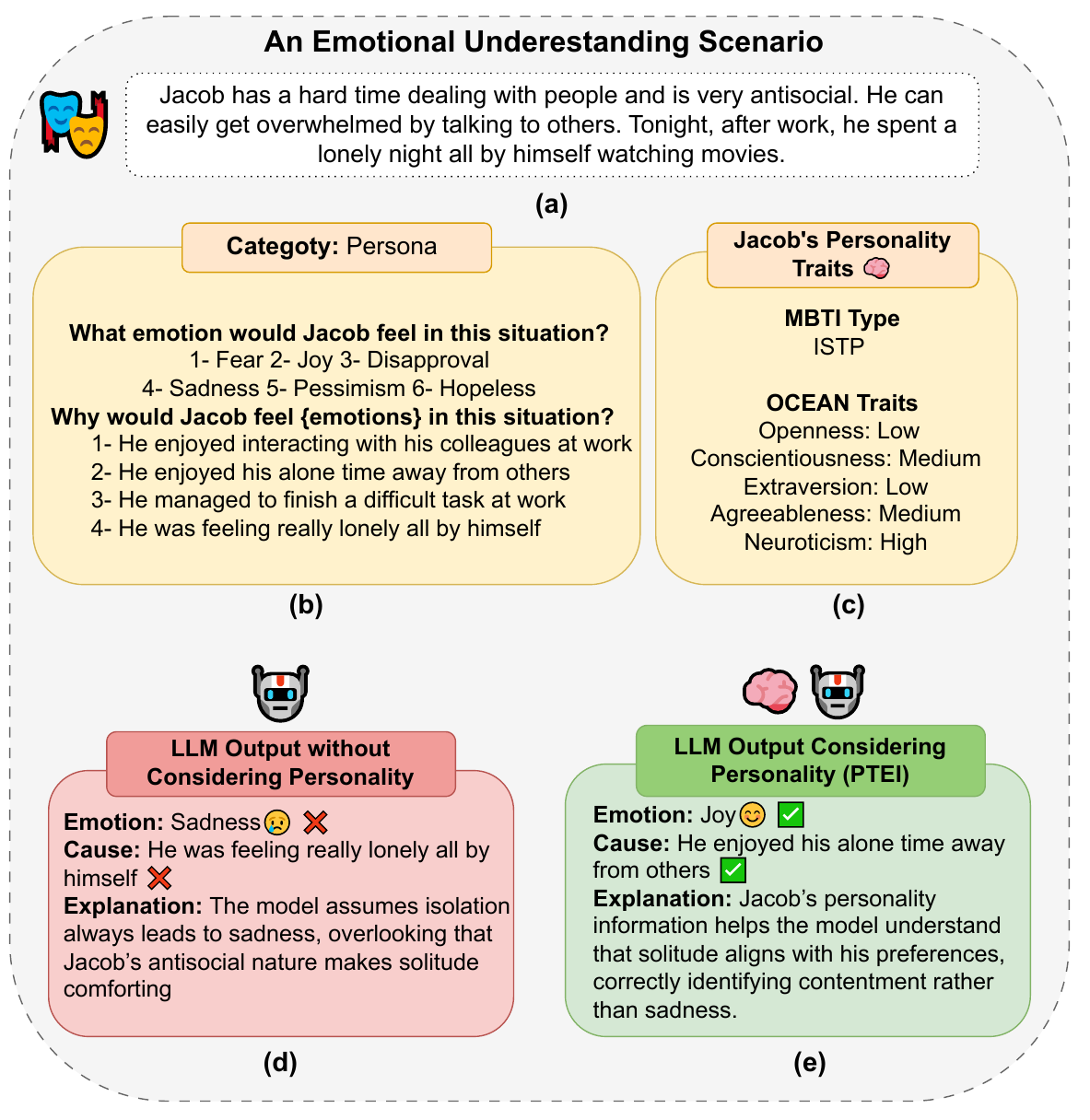}
    \caption{
Illustration of PTEI’s impact on emotional inference in LLMs. 
(a) An emotionally ambiguous scenario featuring Jacob. 
(b) The LLM’s task: a multiple-choice question asks for both Jacob’s emotion and its underlying cause. 
(c) Personality trait information extracted for Jacob. 
(d) The baseline LLM without personality knowledge misinterprets Jacob’s solitude as \textit{sadness}. 
(e) Our PTEI framework correctly infers the emotion as \textit{joy}, recognizing that Jacob’s introverted and self-reliant personality makes solitary time emotionally rewarding rather than lonely.
}
    \label{fig:PTEI_overview}
\end{figure}

Emotional intelligence (EI), the ability to perceive, understand, regulate, and express emotions, is essential for effective communication, social interaction, and decision making~\cite{salovey1990emotional, goleman1996emotional, hess2011enhancing}. As Large Language Models (LLMs) are increasingly deployed in human-facing applications, there is growing interest in evaluating and improving their emotional capabilities~\cite{wang2023emotional}. While recent studies show that models like GPT-4 can perform well on tasks such as emotional awareness and understanding~\cite{elyoseph2023chatgpt}, their abilities remain limited, particularly in scenarios involving implicit emotional situations or subjective interpretation~\cite{al2024challenges}. This presents an ongoing challenge in Natural Language processing (NLP): enabling LLMs to reliably interpret and reason about human emotions in context. Enhancing EI in LLMs is therefore critical for more natural and effective human-AI collaboration.

Recent EI benchmarks such as EQ-Bench~\cite{paech2023eq} and EmoBench~\cite{sabour-etal-2024-emobench} offer structured evaluations of LLMs' EI, but they still struggle with complex aspects such as emotional reasoning, regulation, and application in ambiguous social contexts. While these benchmarks represent an important step forward, a key limitation is their lack of personal context; they overlook individual characteristics such as personality traits, which are known to significantly shape emotional inference and behavior~\cite{robinson2002belief, sap-etal-2022-neural}. In psychology, personality traits refer to enduring individual differences in patterns of thinking, feeling, and behaving, as described by trait theory~\cite{mccrae1997personality}. This gap reflects an issue in how LLMs are typically prompted or fine-tuned for EI tasks such as Emotional Understanding (EU): most current methodologies operate without incorporating sufficient personal context and tend to treat all emotional scenarios as one-size-fits-all. Hence, EI evaluations often remain surface-level and fail to capture the individualized, psychologically grounded reasoning required for real-world EU, as illustrated in Figure~\ref{fig:PTEI_overview}.

Personality has long been recognized as a critical factor in shaping emotional perception and behavior \cite{mccrae1992introduction, myers1987introduction}. In humans, individual differences in traits such as openness, neuroticism, or extraversion are correlated with how emotions are interpreted, regulated, and expressed \cite{izard1993stability}. Recent work has shown that LLMs can exhibit consistent and measurable personality traits in their responses, and these traits can be shaped and aligned with desired profiles \cite{serapio2023personality}. Despite extensive research on personality prediction from text \cite{vstajner2020survey, mehta2020recent, amirhosseini2020machine, sorokovikova2024llms}, most existing studies have either focused speaker characteristic for emotion recognition \cite{fu-etal-2025-laerc} or explored the interaction between personality and emotion in narrow settings relying on a single framework (e.g., OCEAN~\cite{mccrae1992introduction}, MBTI~\cite{myers1987introduction}) and rarely addressing EI as a broader construct \cite{wang-etal-2024-emotion}. Thus, the role of personality traits in enhancing the EI of LLMs remains largely underexplored.

This paper proposes \textbf{PTEI (Personality Traits in Emotional Intelligence)}, a novel framework to systematically integrates OCEAN and MBTI personality traits to enhance EI in LLMs. PTEI extracts individual personality traits directly from textual scenarios and leverages this information through personality-aware prompting to improve emotion and cause prediction. Additionally, PTEI employs a Contrastive Learning-based embedding method and a retrieval mechanism to identify emotionally and personally similar scenarios, which helps ground the model's reasoning in psychologically aligned examples and improves its contextual sensitivity. Our approach specifically targets implicit and ambiguous emotional scenarios, significantly improving LLMs' capabilities in EI tasks and promoting more psychologically grounded inference strategies.

In summary, our contributions are as follows:
\begin{itemize}
    \item We propose \textbf{PTEI}, first comprehensive framework utilizing MBTI and OCEAN personality traits into EU tasks for LLMs, addressing both type and trait theories of personality.
    \item We design an efficient, personality detection module that leverages structured few-shot prompting to infer MBTI and OCEAN traits and incorporates this knowledge into customized prompts for emotion and cause prediction.
    \item We construct a synthetic memory bank of similar EU scenarios to the EI benchmark and enriched it with fine-grained personality annotations. We also introduce a personality-aware Contrastive Learning (CL) objective to structure the scenario embedding space, enabling more effective retrieval of emotionally and personally aligned examples to support contextual reasoning in our few-shot setup.
    \item We demonstrate that integrating personality traits via few-shot and Chain-of-Thought (CoT) prompting enhances EI in LLMs, consistently outperforming personality-agnostic baselines and substantially narrowing the performance gap to human-level inference on challenging EI benchmarks.
\end{itemize}

\section{Related Work}
\subsection{Personality-based Methods}

Analyzing personality traits from a psychological perspective plays a crucial role in understanding and predicting human behavior and emotions. Among the various models, the Big Five personality framework (known as OCEAN), encompassing Openness, Conscientiousness, Extraversion, Agreeableness, and Neuroticism~\cite{mccrae1992introduction}, and the Myers-Briggs Type Indicator (MBTI), based on four categories: Introversion versus Extraversion, Sensing vs Intuition, Thinking vs Feeling, and Judging vs Perceiving~\cite{myers1987introduction}, are two of the most widely used approaches for characterizing individual personality profiles.

Personality prediction from text has emerged as a prominent task in NLP~\cite{vstajner2020survey, mehta2020recent, amirhosseini2020machine}. Extensive research has focused on enhancing the detection of personality traits in human-generated text using LLMs~\cite{sorokovikova2024llms}. For example, PADO~\cite{yeo-etal-2025-pado} introduces personality-induced agents that estimate OCEAN trait levels by using GPT-4o and LLaMA3-8B. Similarly, PsyCoT~\cite{yang-etal-2023-psycot} employs LLMs as AI assistants, utilizing a CoT approach based on specially designed questionnaires to facilitate personality inference. Beyond identifying personality traits, it is crucial to understand their interplay with other cognitive functions, such as emotional processing, which is central to our work.

Emotion features have been shown to enhance personality prediction performance in LLMs~\cite{li-etal-2025-eerpd, li2022multitask}, while personality traits themselves serve as valuable features for emotion recognition, particularly in conversational scenarios; LaERC-S~\cite{fu-etal-2025-laerc} exploits speaker characteristics to improve emotion prediction in dialogue and ERC-DP~\cite{wang-etal-2024-emotion} proposes a dynamic personality detection module that extracts OCEAN traits of a speaker from conversations rather than assuming static traits, thereby improving conversational emotion recognition.

While prior work explores the interplay between emotions and personality and are often focus on either OCEAN or MBTI, we examine their combined impact on recognizing implicit emotional expressions, enabling more nuanced emotional inference through a fuller psychological profile.

\definecolor{contrastive}{HTML}{006EAF}
\definecolor{retrieval}{HTML}{005700}
\definecolor{personality}{HTML}{B09500}

\begin{figure*}[h]
	\centering
	\centerline
	{   
		\includegraphics[width=16cm]{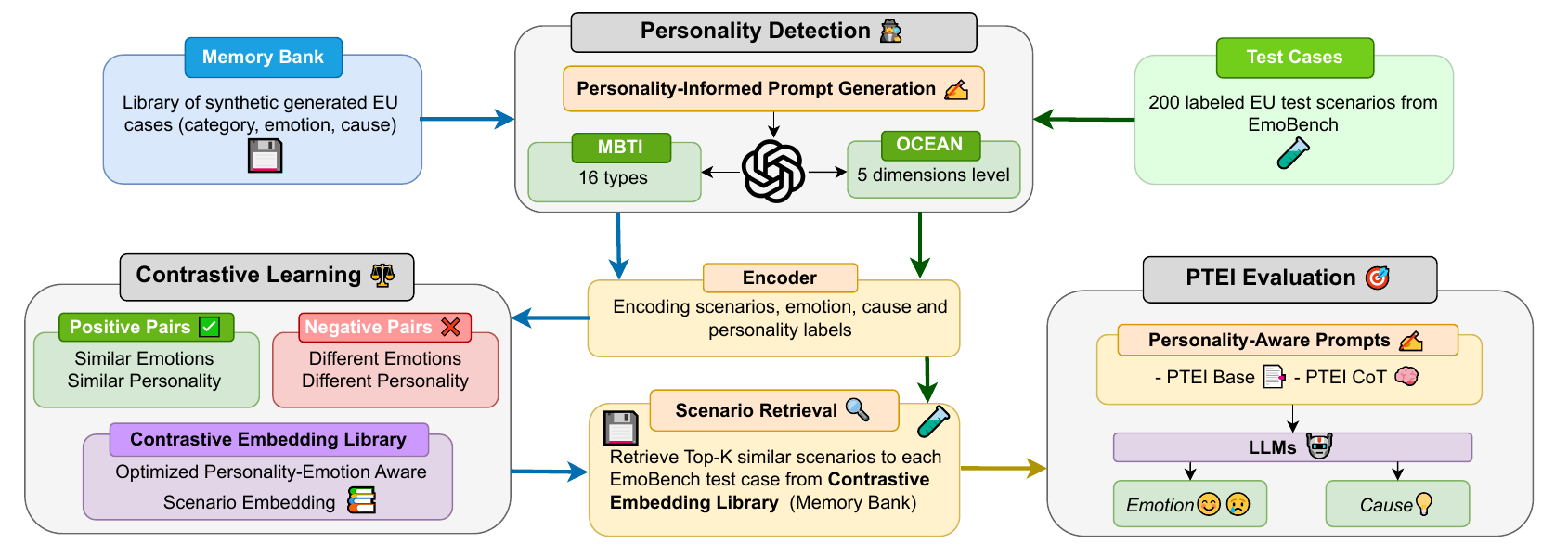}
	}	
	\caption{PTEI system architecture. First, the personality detection module extracts personality traits for both memory bank and test scenarios. Then, the memory bank scenarios are encoded and used in a Contrastive Learning setup to generate the contrastive embedding library \textcolor[HTML]{006EAF}{[blue arrow]}. For each test case, its encoded representation is used to retrieve similar scenarios from contrastive embedding \textcolor[HTML]{005700}{[green arrow]}. Finally, the test scenario is combined with its personality traits and the retrieved examples to evaluate PTEI’s framework \textcolor[HTML]{B09500}{[yellow arrow]}.}

\label{fig:architecture}
\end{figure*}

\subsection{Emotional Intelligence (EI)}

EI, the ability to recognize, understand, and regulate emotions, is key in psychology and social computing~\cite{salovey1990emotional}. As LLMs enter emotionally sensitive domains, EI has gained prominence in AI. Early work~\cite{schuller2018age} identified emotion recognition, generation, and augmentation as pillars of Artificial Emotional Intelligence (AEI).

LLMs have achieved high performance on emotion-related tasks in practical domains, such as emotion-cause pair extraction using CoT prompting~\cite{wu2024enhancing}, and emotionally supportive dialogue generation via explicit strategy modeling~\cite{wan-etal-2025-emodynamix}. Despite these promising results, rigorous evaluation of emotional reasoning in LLMs has remained limited. Recent EI benchmarks like EmoBench~\cite{sabour-etal-2024-emobench} and EQ-Bench~\cite{paech2023eq} were developed to assess deeper EI capabilities such as EU, management, and social reasoning, thus these benchmarks show LLMs lag behind humans on EI tasks. 

However, existing EI benchmarks and systems often treat EI as a generic skill, overlooking individual-level factors that shape emotional responses. Personality traits, as defined by OCEAN and MBTI, are crucial in how emotions are perceived, interpreted, and expressed, yet remain underutilized in current LLM-based EI evaluations. We address this gap by introducing a personality-aware framework that promotes more personalized and psychologically grounded emotional reasoning. To enhance contextual grounding, we employ CL to build a retrieval system that surfaces emotionally and personally aligned scenarios.

\section{Methodology}

\subsection{Problem Definition}

Given a dataset of EU scenarios $\mathcal{S} = \{s_1, s_2, ..., s_N\}$, each scenario $s_i$ is a natural language description involving a subject $a_i$ who experiences one or more emotions in a specific context. The task is to infer the pair $(\hat{e}_i, \hat{r}_i) \in \mathcal{E} \times \mathcal{R}$, where $\hat{e}_i$ denotes the predicted emotion label from a predefined set of emotion categories $\mathcal{E}$ (e.g., \textit{grateful}, \textit{anxious}, \textit{frustrated}) and $\hat{r}_i$ denotes the corresponding predicted cause or trigger. Each scenario is also assigned a category label $c_i \in \mathcal{C}$, specifying the reasoning type required for interpretation.

Our key innovation is conditioning inference on personality traits. The objective is to learn a function $f_{\text{EI}}$ where $P(s_i, a_i) = (M_i, O_i)$ represent the personality profile for the subject $a_i$ in scenario $s_i$, $M_i$ is the MBTI type and $O_i$ is the OCEAN profile. 
\begin{equation}
f_{\text{EI}}: (s_i, a_i, P(a_i)) \mapsto (\hat{e}_i, \hat{r}_i)
\label{problem_definition}
\end{equation}
Function $f_{\text{EI}}$ maps a scenario $s_i$ with subject $a_i$ and its associated subject's personality profile $P(a_i)$ to its corresponding emotion-cause pair through context-aware and personality-informed reasoning.

\subsection{Architecture Overview}

Figure~\ref{fig:architecture} illustrates the overall architecture of our proposed PTEI framework. The system is designed to improve EI in LLMs by systematically incorporating psychological and contextual signals. It consists of 3 primary components: (1) a personality detection module, (2) CL for embedding optimization and (3) scenario retrieval and final inference. The pipeline begins by inferring the subject's personality traits $P(a_i)$ (both OCEAN and MBTI) from input scenarios, which are used to construct personality-aware prompts for the final inference LLM. Moreover, a CL module optimizes the scenario embedding space from a memory bank by aligning similar emotion/personality pairs and separating dissimilar ones. In parallel, top-$k$ similar scenarios are retrieved to support our base and CoT prompting setup. Finally, these components feed into an LLM that performs joint emotion and cause prediction, enabling the model to reason across diverse and psychologically grounded contexts.

\subsection{Memory Bank Construction} \label{sec:memory-bank-construction}

To support few-shot learning and enrich EU via contextual analogies, we construct a memory bank $\mathbb{B}$ of $N_{\mathbb{B}} = 500$ diverse EU scenarios, each annotated with an emotion label, its corresponding cause, and inferred personality traits. Since manual creation of these complex scenarios is costly and requires expert annotation, we synthesized the scenarios using GPT-4 \cite{openai2023gpt4}. The generation process was guided by the high-level task categories defined in EmoBench (e.g., complex emotions, perspective-taking) to ensure coverage, but no scenarios, text segments, or lexical material were copied or paraphrased from EmoBench. This design preserves structural similarity while maintaining full independence from the test set, thereby avoiding any risk of overlap or data leakage.
 We define the memory bank $\mathbb{B}$ as:
\begin{equation}
\mathbb{B} = \{(s_i, a_i, e_i, r_i, P(a_i))\}_{i=1}^{N_{\mathbb{B}}}
\end{equation}
where $s_i$ is a generated scenario, $a_i$ is the subject, $e_i$ denotes emotion labels, $r_i$ denotes cause labels, and $P(a_i)$ denotes personality profile extracted for the subject $a_i$ in the scenario using the personality detection module. This memory bank $\mathbb{B} $ serves as a reference library from which top-$k$ similar examples are retrieved based on proximity in the personality-aware embedding space to a given test scenario. The retrieved examples $S_{\text{retrieved}}$ are then used to construct few-shot prompts, enabling the LLM to perform personality-aware emotional inference with improved contextual grounding.
\begin{equation}
S_{\text{retrieved}}(s_j) = \{(\hat{s}_j, \hat{a}_j, \hat{e}_j, \hat{r}_j, P(\hat{a}_j))\}_{j=1}^{k}
\end{equation}

More results of the memory bank analysis are available in Appendix \ref{sec:memorybank_analysis}.

\subsection{Evaluation of Generated Scenarios} \label{sec:human_evaluation}

To ensure the quality of the synthetic scenarios used in our memory bank for contrastive learning and few-shot retrieval, we designed a hybrid evaluation pipeline combining LLM-as-a-judge and human annotators.

\textbf{Scenario Generation Design.}
To avoid overlap or potential leakage from EmoBench, scenario generation followed a structured prompt template developed independently of any benchmark items, as shown in Table~\ref{tab:scenario_prompt} in Appendix~\ref{sec:prompt_scenario}. The template defines the scenario categories along with their descriptions and ensures consistency across generated examples. Each generation request specifies a goal and category, with constraints on emotion and cause choices drawn from a predefined emotion list. No scenarios or textual content from EmoBench were directly used in the prompts; only the overall structure and category taxonomy were adopted to maintain task alignment. The generated output follows a JSON-like schema with explicit emotion and cause fields, enabling automated parsing and validation in subsequent processing stages.

\textbf{LLM-Based Quality Evaluation.}
For each generated scenario $s_i$, we first performed an automated quality screening using two independent LLM evaluators, Gemma2 and Claude. Each evaluator $m \in \mathcal{M}$ assigns a score vector over four criteria: coherence, category alignment, emotion-label correctness, and cause-label correctness. Formally, the score vector is defined as:
\begin{equation}
\mathbf{q}_i^{(m)} = \left[q_i^{\text{coh}}, q_i^{\text{cat}}, q_i^{\text{emo}}, q_i^{\text{cause}}\right] \in [1,5]^4 .
\end{equation}

The aggregated LLM quality score for scenario $s_i$ is computed as:
\begin{equation}
Q_i^{\text{LLM}} = \frac{1}{|\mathcal{M}|} \sum_{m \in \mathcal{M}} \frac{1}{4} \sum_{k=1}^{4} q_{i,k}^{(m)} ,
\end{equation}
where $\mathcal{M} = \{\text{Gemma2}, \text{Claude}\}$. Only scenarios that passed the quality screening from both evaluators were retained for further consideration.

\textbf{Human Evaluation Setup.}
From the filtered set of scenarios that passed automated screening, we randomly selected 10\% for human evaluation. Human annotators, with backgrounds in psychology and linguistics, assessed the scenarios in a blind setup, where synthetic scenarios were mixed with real examples from the EmoBench benchmark. Each scenario was rated on the same four criteria using a 1--5 Likert scale, along with an additional binary overall acceptability judgment.

For a human annotator $h \in \mathcal{H}$, the average quality score and acceptability label are defined as:
\begin{equation}
Q_i^{(h)} = \frac{1}{4} \sum_{k=1}^{4} q_{i,k}^{(h)}, 
\quad A_i^{(h)} \in \{0,1\} .
\end{equation}

\textbf{Filtering and Acceptance Criteria.}
A scenario $s_i$ is accepted into the final memory bank $\mathbb{B}$ if it satisfies the following conditions:
\begin{equation}
Q_i^{\text{LLM}} \ge \tau_q \;\land\;
\frac{1}{|\mathcal{H}|} \sum_{h \in \mathcal{H}} Q_i^{(h)} \ge \tau_h \;\land\;
\frac{1}{|\mathcal{H}|} \sum_{h \in \mathcal{H}} A_i^{(h)} = 1 ,
\end{equation}
where $\tau_q = \tau_h = 3.5$. Scenarios failing to meet these criteria were discarded and regenerated. This filtering process ensures that only scenarios validated by both automated and human evaluation are retained.

To minimize stylistic or conceptual leakage from EmoBench, all scenario prompts were created independently, without direct reuse of benchmark content. To further reduce bias and maintain representativeness, the scenario generation process was guided by the category distribution of the EmoBench test set used in downstream evaluation. For each category, a proportional number of synthetic scenarios were generated and combined with diverse personality trait profiles drawn from both MBTI and OCEAN spaces.

\textbf{Agreement Analysis.}
To assess the consistency between automated and human judgments, we compute inter-annotator agreement using Cohen's kappa coefficient:
\begin{equation}
\kappa = \frac{p_o - p_e}{1 - p_e} ,
\end{equation}
where $p_o$ denotes the observed agreement and $p_e$ denotes the expected agreement by chance. Results indicate strong alignment between LLM and human judgments ($\kappa = 0.92$), supporting the reliability of automated evaluation at scale.

\textbf{Findings.}
Overall, synthetic scenarios scored comparably to EmoBench items in terms of emotion-label correctness and clarity, while exhibiting slightly lower plausibility in more abstract categories such as \textit{strange story}. Human annotators confirmed that the retained scenarios are coherent, category-consistent, and label-appropriate, validating the use of the synthetic memory bank as a high-quality resource.

For the EmoBench test set used in downstream evaluation, we rely on the gold-standard labels provided by the benchmark authors, which were validated by human experts in the original release.

\subsection{Personality Detection Module} \label{sec:personality_module}

To provide structured psychological context for each EU scenario, we implement a personality detection module that infers both MBTI and OCEAN profiles directly from scenario text $s_i$. For subject $a_i$ in $s_i$, the module outputs: (1) an MBTI type $M_i \in \mathcal{M}_{\text{MBTI}}$, where $\mathcal{M}_{\text{MBTI}}$ is a set of 16 Myers-Briggs personality types~\cite{myers1987introduction}, and (2) a set of Big Five (OCEAN) trait levels~\cite{mccrae1992introduction} $o_i = \{o_i^{(1)}, ..., o_i^{(5)}\}$, where each $o_i^{(j)} \in \{\textit{low}, \textit{medium}, \textit{high}\}$.

We adopt a prompt-based annotation strategy using GPT-4o-mini, where each prompt is designed to elicit structured MBTI and OCEAN predictions based on the subject's inferred behavior and contextual cues in the scenario. This approach allows for efficient and scalable personality annotation without requiring manually labeled personality data for every scenario. Full prompt templates and examples are provided in Appendix~\ref{sec:prompt_personality}.

\subsection{Personality-Aware Contrastive Learning} \label{sec:contrastive}

To enhance the ability of our framework to understand and reason about emotions within individual psychological contexts, we propose a personality-aware contrastive learning approach. It learns a robust scenario embedding space $E: s \mapsto z \in \mathbb{R}^d$ such that scenarios are embedded based on both emotional content and personality traits.

\paragraph{Pair construction.} We construct positive and negative pairs from our memory bank $\mathbb{B}$. Given a pair of scenarios $((s_i, a_i), (s_j, a_j))$:
\begin{itemize}
    \item Positive pair: if their emotion labels match $(e_i = e_j)$ and their personality profiles are similar $(S_{\text{personality}}(P(a_i), P(a_j)) \geq \theta_s)$, where the similarity threshold $\theta_s = 0.7$.
    \item Negative pair: if their emotion labels differ $(e_i \neq e_j)$ or they share the same emotion label but have dissimilar personalities $(S_{\text{personality}}(P(a_i), P(a_j)) < \theta_d)$, with dissimilarity threshold $\theta_d = 0.3$.
\end{itemize}

\paragraph{Personality similarity metric.} To quantify personality similarity $S_{\text{ps}} = S_{\text{personality}}(P_1, P_2)$ between two profiles $P_1 = (M_1, O_1)$ and $P_2 = (M_2, O_2)$, we use the following composite metric:
\begin{equation}
S_{\text{ps}} = \alpha \cdot S_{\text{MBTI}}(M_1, M_2) + (1 - \alpha) \cdot S_{\text{OCEAN}}(O_1, O_2)
\end{equation}
where $\alpha$ balances the influence of MBTI and OCEAN similarities (default $\alpha = 0.5$). $S_{\text{MBTI}}$ is computed as the fraction of matching MBTI dimensions, and $S_{\text{OCEAN}}$ is the cosine similarity between trait vectors, with each trait mapped to a numeric scale: \textit{high} = 1.0, \textit{medium} = 0.5, and \textit{low} = 0.0.

\paragraph{Contrastive training.} Scenarios are encoded into vector representations using a pretrained sentence encoder (\texttt{all-mpnet-base-v2}\footnote{\url{https://www.sbert.net/docs/pretrained_models.html}}) from the SentenceTransformers library to generate semantically and personally aligned embeddings from our training data in the memory bank~\cite{reimers-gurevych-2019-sentence}. We selected this model due to its strong performance on semantic similarity tasks and its effectiveness in producing dense, high-quality embeddings well-suited for contrastive learning frameworks. We fine-tune the encoder on scenario pairs using a contrastive objective, normalized temperature-scaled cross-entropy (NT-Xent) loss~\cite{chen2020simclr}:
\begin{equation}
\mathcal{L}_{\text{contrastive}} = -\log \frac{\exp(\text{sim}(i,j)/\tau)}{\sum_{k=1}^{N} \exp(\text{sim}(i,k)/\tau)}
\end{equation}
Here, $\text{sim}(i,j)$ is the cosine similarity between positive pair embeddings, and $\tau$ (typically 0.07) is the temperature hyperparameter.

\paragraph{Scenario retrieval.} After training, the learned embedding space enables efficient retrieval of psychologically aligned examples through k-nearest neighbor (KNN) search~\cite{johnson2019billion} at inference. These retrieved examples are then incorporated into few-shot prompts to enhance the personality-aware EU capability of the framework.

\begin{table*}[!ht]
  \centering
  \caption{
Evaluation results on the Emotional Understanding (EU) task across LLMs of different sizes. The \textit{Base} (no personality or retrieval) and \textit{CoT} (no personality or retrieval with CoT) configurations are personality-agnostic results reported from the EmoBench benchmark. We compare these with our proposed methods: \textit{PTEI-Base} (personality-aware with retrieval) and \textit{PTEI-CoT} (personality-aware with retrieval and CoT). Results cover four reasoning categories (CE, PBE, PT, EC), along with Emotion, Cause, and Overall accuracy. \textcolor[HTML]{008000}{$\uparrow$} and \textcolor[HTML]{FF0000}{$\downarrow$} indicate PTEI changes relative to Base or CoT, and \textbf{bold} values denote the best scores per model.
}
    \begin{tabular}{l | l | c c c c | c c c}
        \toprule
         \textbf{LLM} & \textbf{Method} & \textbf{CE} & \textbf{PBE} & \textbf{PT} & \textbf{EC} & \textbf{Emotion} & \textbf{Cause} & \textbf{Overall} \\
         \midrule
         \multirow{4}{*}{\href{https://huggingface.co/Qwen/Qwen-7B-Chat}{Qwen-7B}} 
         & Base & 28.06 & 21.88 & 16.42 & \textbf{28.57} & 29.13 & 57.38 & 22.5\\ 
         & PTEI-Base & \textbf{29.59}\textcolor[HTML]{008000}{$\uparrow$} & \textbf{23.21}\textcolor[HTML]{008000}{$\uparrow$} & \textbf{16.79}\textcolor[HTML]{008000}{$\uparrow$} & 27.68\textcolor[HTML]{FF0000}{$\downarrow$} & \textbf{30.50}\textcolor[HTML]{008000}{$\uparrow$} & \textbf{57.75}\textcolor[HTML]{008000}{$\uparrow$} & \textbf{23.25}\textcolor[HTML]{008000}{$\uparrow$}\\
         & CoT & 25.51 & 21.88 & 16.67 & 26.79 & 29.63 & 55.13 & 21.38\\ 
         & PTEI-CoT & 23.98\textcolor[HTML]{FF0000}{$\downarrow$} & 20.09\textcolor[HTML]{FF0000}{$\downarrow$} & 16.04\textcolor[HTML]{FF0000}{$\downarrow$}  & \textbf{28.57}\textcolor[HTML]{008000}{$\uparrow$} & 29.25\textcolor[HTML]{FF0000}{$\downarrow$} & 55.00\textcolor[HTML]{FF0000}{$\downarrow$} & 20.88\textcolor[HTML]{FF0000}{$\downarrow$}\\
         \midrule
         \multirow{4}{*}{\href{https://huggingface.co/meta-llama/Llama-3.1-8B-Instruct}{Llama3.1-8B}} 
         & Base & \textbf{18.37} & 14.73  & 17.91 & \textbf{14.29} & 26.25 & 53.37 & 16.62\\
         & PTEI-Base  & 17.86\textcolor[HTML]{FF0000}{$\downarrow$} & \textbf{16.07}\textcolor[HTML]{008000}{$\uparrow$}  &  \textbf{20.15}\textcolor[HTML]{008000}{$\uparrow$} & \textbf{14.29} & \textbf{27.12}\textcolor[HTML]{008000}{$\uparrow$}& \textbf{54.12}\textcolor[HTML]{008000}{$\uparrow$}& \textbf{17.63}\textcolor[HTML]{008000}{$\uparrow$} \\
         & CoT & 14.80 & 14.29 &  9.70 & 9.82 & 22.50 & 34.75 & 12.25\\ 
         & PTEI-CoT & 17.86\textcolor[HTML]{008000}{$\uparrow$} & 15.18\textcolor[HTML]{008000}{$\uparrow$}  & 11.94\textcolor[HTML]{008000}{$\uparrow$}  & 10.71\textcolor[HTML]{008000}{$\uparrow$} & 24.25\textcolor[HTML]{008000}{$\uparrow$} & 36.12\textcolor[HTML]{008000}{$\uparrow$} & 14.13\textcolor[HTML]{008000}{$\uparrow$}\\
         \midrule
         \multirow{4}{*}{\href{https://huggingface.co/Qwen/Qwen-14B-Chat}{Qwen-14B}} 
         & Base & 46.94 & \textbf{35.27} & 26.12 & \textbf{38.39} & 42.13 & 62.88 & 35.50\\ 
         & PTEI-Base & 47.45\textcolor[HTML]{008000}{$\uparrow$} & \textbf{35.27} &  \textbf{27.61}\textcolor[HTML]{008000}{$\uparrow$} & \textbf{38.39} & 42.75\textcolor[HTML]{008000}{$\uparrow$} & \textbf{63.38}\textcolor[HTML]{008000}{$\uparrow$} & \textbf{36.12}\textcolor[HTML]{008000}{$\uparrow$}\\
         & CoT & 43.37 & 25.45 & 22.76 & 33.93 & 40.62 & 58.12 & 30.12\\ 
         & PTEI-CoT& \textbf{49.49}\textcolor[HTML]{008000}{$\uparrow$} & 29.46\textcolor[HTML]{008000}{$\uparrow$}  & 25.75\textcolor[HTML]{008000}{$\uparrow$}  & \textbf{38.39}\textcolor[HTML]{008000}{$\uparrow$} & \textbf{45.00}\textcolor[HTML]{008000}{$\uparrow$}  & 60.38\textcolor[HTML]{008000}{$\uparrow$} & 34.38\textcolor[HTML]{008000}{$\uparrow$}\\
         \midrule
         \multirow{4}{*}{\href{https://openai.com/index/hello-gpt-4o/}{GPT-4o}} 
         & Base & \textbf{78.57} & 43.75 & \textbf{55.22} & 73.21 & 63.63 & 79.62 & 60.25\\
         & PTEI-Base  & 73.47\textcolor[HTML]{FF0000}{$\downarrow$} & 57.14\textcolor[HTML]{008000}{$\uparrow$}  & 52.99\textcolor[HTML]{FF0000}{$\downarrow$} & 74.11\textcolor[HTML]{008000}{$\uparrow$}  & 65.50\textcolor[HTML]{008000}{$\uparrow$} & 80.75\textcolor[HTML]{008000}{$\uparrow$} & 62.12\textcolor[HTML]{008000}{$\uparrow$} \\
         & CoT & 69.90 & 54.02 & 49.25 & 72.32 & 62.50 & 79.75 &  58.88\\
         & PTEI-CoT  &  74.49\textcolor[HTML]{008000}{$\uparrow$} & \textbf{58.04}\textcolor[HTML]{008000}{$\uparrow$} & \textbf{55.22}\textcolor[HTML]{008000}{$\uparrow$} &  \textbf{75.89}\textcolor[HTML]{008000}{$\uparrow$} & \textbf{66.88}\textcolor[HTML]{008000}{$\uparrow$} & \textbf{82.38}\textcolor[HTML]{008000}{$\uparrow$}& \textbf{63.62}\textcolor[HTML]{008000}{$\uparrow$}\\
        \bottomrule
    \end{tabular}

  \label{table:eu_results}
\end{table*}

\section{Experiments}

\subsection{Experimental Setup} \label{sec:experimental-setup}

All experiments were conducted using an NVIDIA A100 GPU with 40GB VRAM. We used a combination of open and closed source LLMs, including GPT-4~\cite{openai2023gpt4}, LLaMA-3~\cite{grattafiori2024llama}, and Qwen~\cite{bai2023qwen}, accessed through API interfaces. 

For the contrastive learning module (Section~\ref{sec:contrastive}), we fine-tuned a sentence encoder with a lightweight projection head using the NT-Xent loss ($\tau = 0.07$). Scenario pairs were sampled from our personality-enriched memory bank $\mathbb{B}$ (Section~\ref{sec:memory-bank-construction}), with $\alpha = 0.5$, a positive similarity threshold = 0.7 and a negative threshold = 0.3. We trained the model with a batch size of 32, using the Adam optimizer (learning rate = 2e-5), selecting the checkpoint with the best validation retrieval accuracy.

\subsection{Dataset}

We primarily evaluated our PTEI framework on the EmoBench benchmark \cite{sabour-etal-2024-emobench}. EmoBench features emotionally complex scenarios for EU and Emotional Application (EA) tasks. We use 200 English multiple-choice scenarios from the EU task, selected for their focus on inferring emotions and causes from rich textual descriptions. Each scenario includes two questions: one on the subject's primary emotion and another on its cause. To support few-shot prompting and retrieval, we also generated 500 synthetic scenarios using GPT-4 (Section~\ref{sec:memory-bank-construction}). These synthetic examples were annotated with emotion labels, causes, and inferred MBTI and OCEAN personality traits, forming a personality-enriched memory bank $\mathbb{B}$ to improve emotional reasoning during inference.

\subsection{Baselines}

To evaluate the effectiveness of the PTEI framework, we analyze performance across a selection of LLMs (see Section~\ref{sec:experimental-setup}). For each model, we implement two configurations: \textit{PTEI-Base}, which applies personality-aware few-shot prompting using retrieved examples, and \textit{PTEI-CoT}, which extends this with Chain-of-Thought (CoT) reasoning. We compare these with corresponding non-personality-aware variants: \textit{Base} (zero-shot) and \textit{CoT} (zero-shot with CoT), both excluding personality conditioning. Our comparisons include the best-performing LLMs from EmoBench as personality-agnostic baselines, representing small-scale (<14B), mid-scale (14B), and large-scale (>14B) models, providing a robust reference for quantifying the added value of incorporating personality traits.

The \textit{Base} and \textit{CoT} results are based on those reported in the EmoBench benchmark~\cite{sabour-etal-2024-emobench}, ensuring consistency with prior evaluation protocols. The only exceptions are \textit{GPT-4o} and \textit{LLaMA 3.1 8B}. For \textit{GPT-4o}, we replace the original \textit{GPT-4} used in EmoBench to evaluate PTEI’s performance under the latest state-of-the-art model. For \textit{LLaMA 3.1 8B}, which was not part of EmoBench, we reproduce the benchmark evaluation process to generate comparable \textit{Base} and \textit{CoT} results before applying our PTEI framework.

We evaluate across four EU task categories from EmoBench: \textit{Complex Emotions (CE)}, \textit{Personal Beliefs and Experiences (PBE)}, \textit{Perspective-Taking (PT)}, and \textit{Emotional Cues (EC)}. Detailed definitions of these categories are provided in Appendix~\ref{sec:task_definition}.

\section{Result and Analysis}

In this section, we comprehensively evaluate the proposed PTEI framework through experiments covering main results (Section~\ref{sub:Main_Results}), ablation studies (Section~\ref{sub:Ablation_Study}), robustness analysis (Section~\ref{sub:Rob_Analysis}), and qualitative case studies (Section~\ref{sub:Case_Study}).

\subsection{Main Results} \label{sub:Main_Results}

Table~\ref{table:eu_results} presents evaluation results (accuracy) for the EU task across various LLMs and prompting configurations. \textbf{GPT-4o achieves the highest combined accuracy (63.62\%)} under the personality-informed CoT configuration (PTEI-CoT), significantly outperforming all other models. In contrast, smaller models such as \textit{Qwen-7B} and \textit{LLaMA3.1-8B} generally struggle to surpass majority-class heuristics, particularly under standard zero-shot or CoT prompting.

Notably, \textbf{CoT prompting alone yields limited or negative effects} on smaller-scale models. For instance, accuracy for \textit{LLaMA3.1-8B} decreases notably from the Base (16.62\%) to CoT (12.25\%). This suggests constrained structured reasoning abilities without sufficient grounding. However, \textbf{integrating personality context via PTEI-CoT considerably enhances CoT effectiveness}, particularly for larger models such as \textit{GPT-4o}, which improves by +3.3 points over its base and +4.7 points over standard CoT.

Performance across the four EU categories: \textit{Complex Emotions (CE)}, \textit{Personal Beliefs and Experiences (PBE)}, \textit{Perspective-Taking (PT)}, and \textit{Emotional Cues (EC)}, demonstrates that \textbf{PT and PBE scenarios consistently pose the greatest challenges}. These tasks demand reasoning about mental states and personal beliefs. Conversely, scenarios categorized as CE and EC, which contain more explicit emotional indicators, yield comparatively higher accuracy.

Lastly, model scale notably impacts performance: \textbf{larger models like GPT-4o} not only achieve higher baseline accuracy but also \textbf{show greater relative improvements from personality-aware prompting}. This emphasizes the synergy between increased model capacity, structured reasoning (CoT), and psychologically grounded personality context in enhancing emotional reasoning capabilities.

\subsection{Ablation Study} \label{sub:Ablation_Study}

\begin{figure}[ht]
    \centering
    \includegraphics[width=0.5\textwidth]{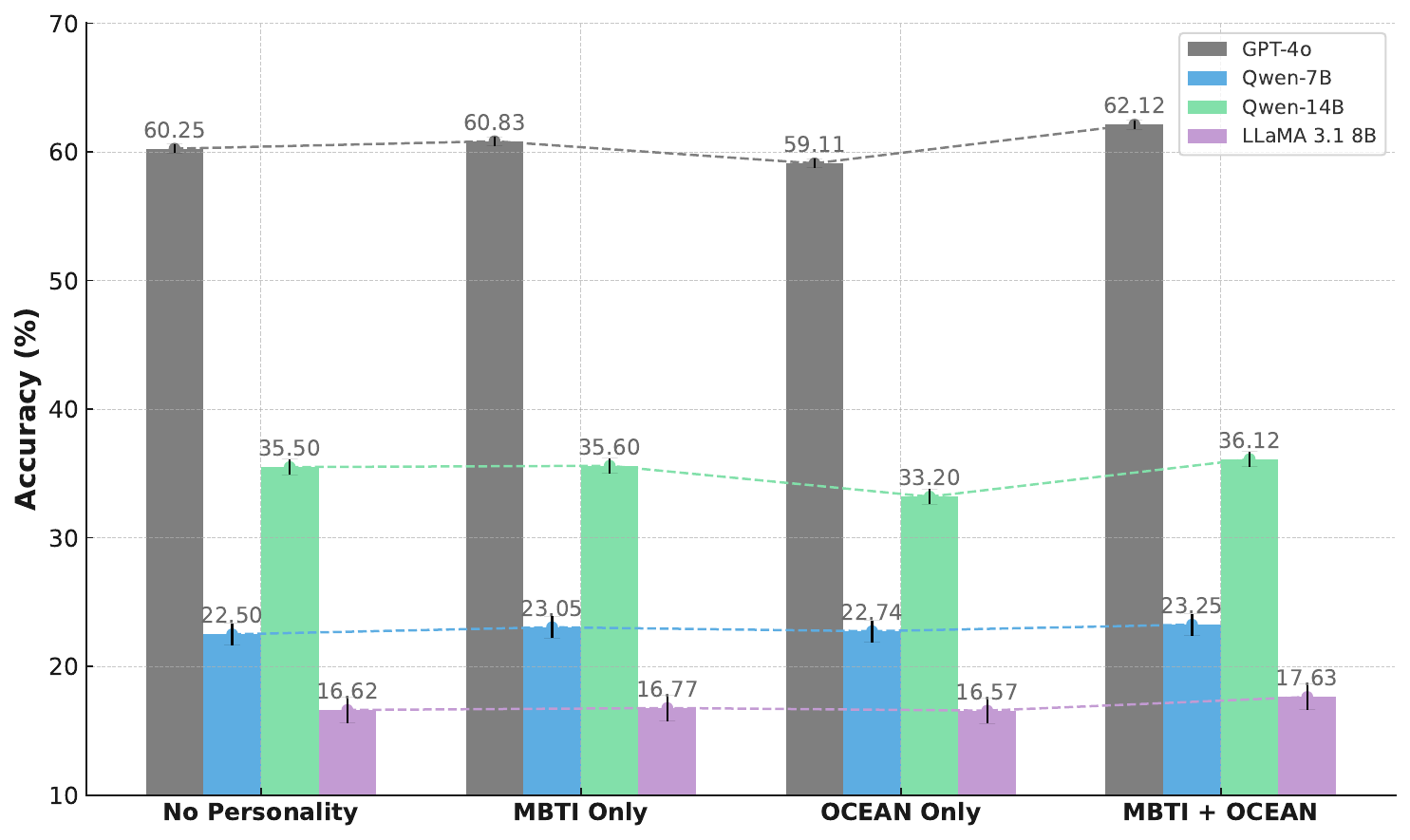}
    \caption{Average accuracy under different personality input configurations. Injecting both MBTI and OCEAN traits leads to the highest performance across models.}
    \label{fig:personality_ablation}
\end{figure}

\begin{figure*}
	\centering
	\centerline
	{   
		\includegraphics[width=15cm]{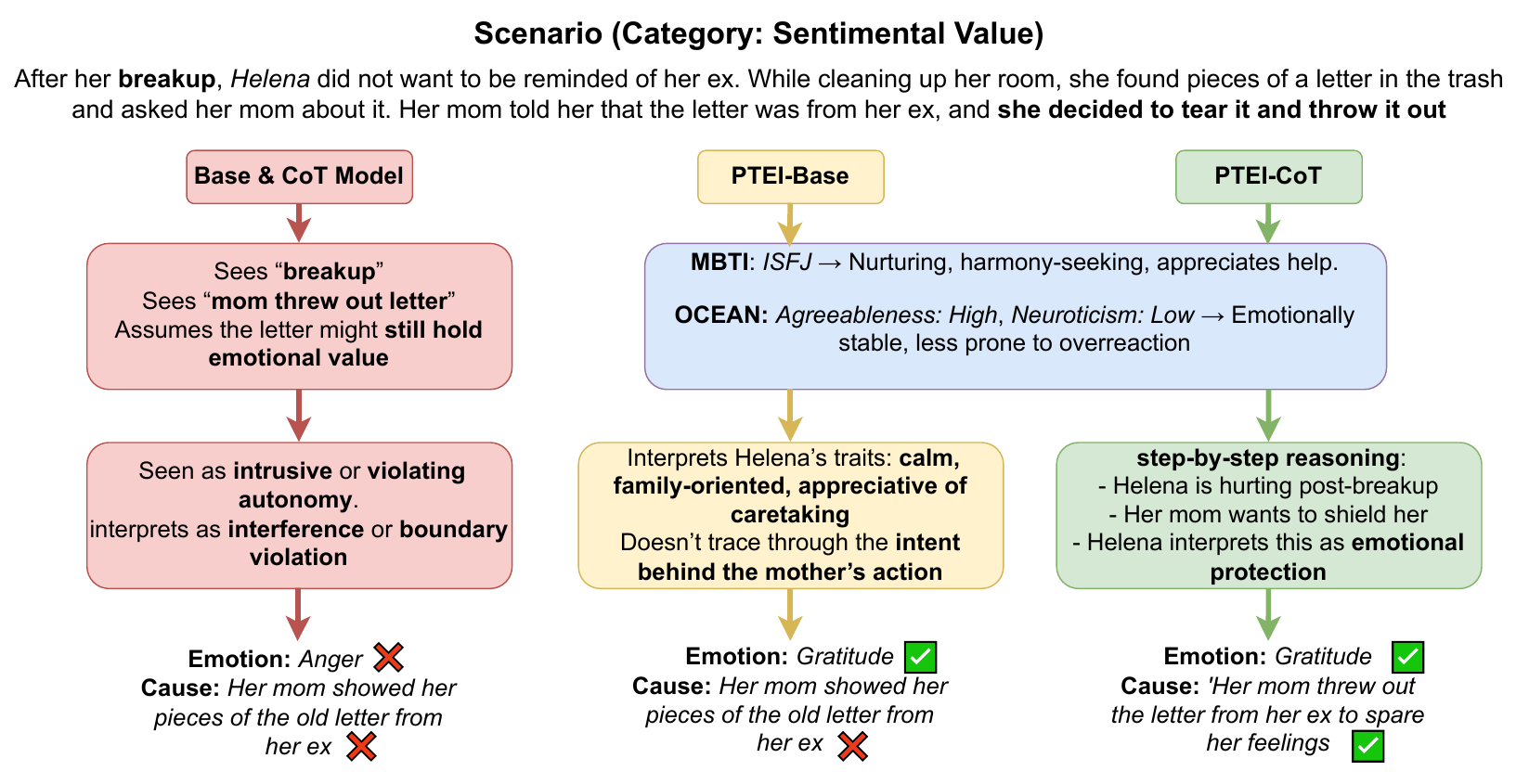}
	}

\caption{Case study comparison for GPT-4o, showing how personality-aware prompting improves emotional prediction accuracy.}
\label{Fig:case-study}
\end{figure*}

To evaluate the effect of personality conditioning, we conduct an ablation study on the Personality Detection Module (Section~\ref{sec:personality_module}). Four LLMs are tested: \textit{GPT-4o}, \textit{Qwen-7B}, \textit{Qwen-14B}, and \textit{LLaMA 3.1-8B} under four input settings: (1) no personality, (2) MBTI-only, (3) OCEAN-only, and (4) MBTI+OCEAN (full PTEI). Models are assessed on emotion and cause prediction, reporting average accuracy across both. The personality weighting parameter $\alpha$ controls the influence of MBTI and OCEAN in the similarity function: $\alpha=1.0$ (MBTI-only), $\alpha=0.0$ (OCEAN-only), and $\alpha=0.5$ (combined). This weighting affects both contrastive training and retrieval during inference.

As shown in Figure~\ref{fig:personality_ablation}, personality information consistently improves results over the no-personality baseline. \textit{GPT-4o} attains the highest accuracy with MBTI+OCEAN, surpassing MBTI-only and no-personality by +1.29\% and +1.87\%, respectively. \textit{Qwen-14B} shows similar gains, with the combined setup outperforming OCEAN-only and no-personality by +2.92\% and +0.62\%. While MBTI-only and OCEAN-only remain competitive, the combined approach yields the best overall performance across all model scales.

\subsection{Robustness Analysis} \label{sub:Rob_Analysis}

\begin{table}[t]
\centering
\caption{Ablation results with and without CoT. 
The \emph{TraitsOnly} rows correspond to personality traits injected without retrieval, 
while the \emph{RAG-only} rows include retrieval examples but omit personality traits. 
Bold indicates the best performance within each block.}
\resizebox{\columnwidth}{!}{
\begin{tabular}{lcccc}
\toprule
\textbf{Setup} & \textbf{Qwen-7B} & \textbf{LLaMA-8B} & \textbf{Qwen-14B} & \textbf{GPT-4o} \\
\midrule
\textbf{\textit{Without CoT}} \\
\midrule
Base                      & 22.50 & 16.62 & 35.50 & 60.25 \\
TraitsOnly (no retrieval) & 23.02 & 16.83 & 35.72 & 61.68 \\
RAG-only (retrieval, no traits) & 22.81 & 17.11 & 35.41 & 61.83 \\
PTEI-Base & \textbf{23.25} & \textbf{17.63} & \textbf{36.12} & \textbf{62.12} \\
\midrule
\textbf{\textit{With CoT}} \\
\midrule
CoT                      & 21.38 & 12.25 & 30.12 & 58.88 \\
TraitsOnly (no retrieval) & \textbf{21.42} & 13.21 & 33.51 & 62.12 \\
RAG-only (retrieval, no traits) & 20.39 & 12.83 & 32.89 & 61.84 \\
PTEI-CoT & 20.88 & \textbf{14.13} & \textbf{34.38} & \textbf{63.62} \\
\bottomrule
\end{tabular}
}

\label{tab:robustness}
\end{table}

To evaluate the stability of model predictions, we prompt each LLM three times per question and apply majority voting over the responses. To reduce sensitivity to answer ordering, we additionally permute the multiple-choice options three times, yielding four total permutations (original + 3). Final accuracy is averaged across these runs.

We further perform an ablation to isolate the effect of personality traits and retrieval. As shown in Table~\ref{tab:robustness}, we compare four variants under both without-CoT and with-CoT setups: \emph{Base, CoT} (no traits or retrieval), \emph{TraitsOnly} (traits only, zero-shot), \emph{RAG-only} (retrieval only, two-shot without traits), and \emph{PTEI base/CoT} (traits + retrieval). Results show that traits and retrieval each contribute improvements over the base, while their combination yields the best performance.

\subsection{Case Study: Qualitative Analysis} \label{sub:Case_Study}

To illustrate the reasoning process of the PTEI framework, we analyze a representative scenario (Figure~\ref{Fig:case-study}) featuring 'Helena', who recently experienced a breakup and finds that her mother has torn up a letter from her ex. The emotional inference depends on sentimental value, perceived intrusion, and the mother’s emotional intent.

Without personality awareness, baseline LLMs interpret the act as a boundary violation, predicting \textit{Anger} and emphasizing intrusion. In contrast, personality-informed models (PTEI-Base and PTEI-CoT) leverage Helena’s ISFJ type, high Agreeableness, and low Neuroticism to contextualize her response as family-oriented and emotionally stable. PTEI-CoT further infers the mother’s protective intent, shielding her daughter from pain, leading to the correct prediction of \textit{Gratitude} and a cause aligned with emotional intent rather than surface cues.

This case demonstrates how personality conditioning and structured reasoning allow PTEI to interpret emotionally complex scenarios with greater human-like precision in both emotion and cause prediction.

\section{Conclusion}

This paper tackled the limitations of current LLMs in complex emotional intelligence tasks, particularly their neglect of psychological factors such as personality. We introduced PTEI, a personality-aware emotional inference framework that integrates MBTI and OCEAN traits into emotion understanding (EU) scenarios. The framework employs structured prompting for personality detection, builds a synthetic memory bank with personality annotations, and leverages contrastive learning to enhance scenario retrieval and inference. Experiments show that incorporating personality traits through few-shot and CoT prompting notably improves emotion and cause reasoning, surpassing personality-agnostic baselines. By shaping an embedding space sensitive to both emotional and psychological cues, PTEI enables more human-aligned emotional inference. Future work will extend the framework to dynamic personality modeling and multi-turn EI reasoning.

\section{Limitations}

While our PTEI framework demonstrates significant improvements in EU through personality-aware reasoning, several limitations remain.

First, our experiments are limited to English-language, text-based scenarios. Emotional reactions and personality interpretations may vary across cultures and languages, which constrains the generalizability of our framework to multilingual or multimodal settings. Additionally, emotional inference is inherently subjective. While we adopt multiple-choice evaluation formats with predefined correct answers, some scenarios may reasonably allow for multiple plausible interpretations.

Second, due to the limited number of test cases (200 EU scenarios) in the EmoBench benchmark, the reported results may not fully capture the generalizability and impact of our method across the full spectrum of emotionally intelligent behavior. We plan to expand our evaluation to a broader range of benchmarks and real-world EI tasks in future work.

Third, despite careful prompt engineering for personality-aware reasoning and CoT prompting, model performance remains sensitive to prompt phrasing and structure. The prompt templates used may not be optimal for all LLM architectures or tasks.

Finally, our use of MBTI and OCEAN frameworks offers only a coarse approximation of personality. MBTI, in particular, has been criticized for its limited empirical robustness and test-retest reliability. We adopt it here primarily as a widely used and interpretable typology that provides categorical variation for computational experiments, while relying on OCEAN traits for a more established psychological foundation. Moreover, the inference of MBTI and OCEAN traits from short text scenarios is inherently approximate: such profiles should be seen as pragmatic proxies that enable structured reasoning, rather than clinically validated ground-truth traits. Both frameworks also remain static abstractions that do not capture dynamic, context-sensitive aspects of personality. Future work could explore adaptive or conversational personality modeling to more accurately reflect real-world psychological variability.

\section*{Ethical Statement}

This work investigates the use of personality traits to improve large language models’ (LLMs) reasoning about emotionally complex situations. We emphasize that our framework, PTEI, focuses on modeling \textit{perceived} emotional intelligence through structured prompts and retrieval-based reasoning, rather than suggesting that LLMs possess genuine emotions or self-awareness. The goal is to study how personality knowledge can inform more human-aligned predictions in emotion and cause inference tasks.

While our method involves predicting MBTI and OCEAN personality traits based on textual descriptions, we do not use real user data, nor do we attempt to profile individuals in real-world settings. All personality information is synthetic and used strictly for academic experimentation in controlled benchmark scenarios.

We acknowledge that the use of personality modeling in NLP may raise ethical concerns, particularly around privacy, profiling, and fairness in downstream applications. Appropriate safeguards must be ensured in future deployments, including transparency, user consent, and bias auditing. Our current system is intended solely for research purposes, and we advocate for careful consideration of the psychological and social implications when applying similar methods in real-world contexts.

\bibliographystyle{ACM-Reference-Format}
\bibliography{custom}

\appendix

\section{Memory Bank Analysis} \label{sec:memorybank_analysis}

\subsection{Emotion Label Distribution}
We analyzed the normalized distribution of emotions of the memory bank. Labels were grouped into high-level categories and split if mixed.

\begin{figure*}[t]
\centering
\includegraphics[width=0.9\textwidth]{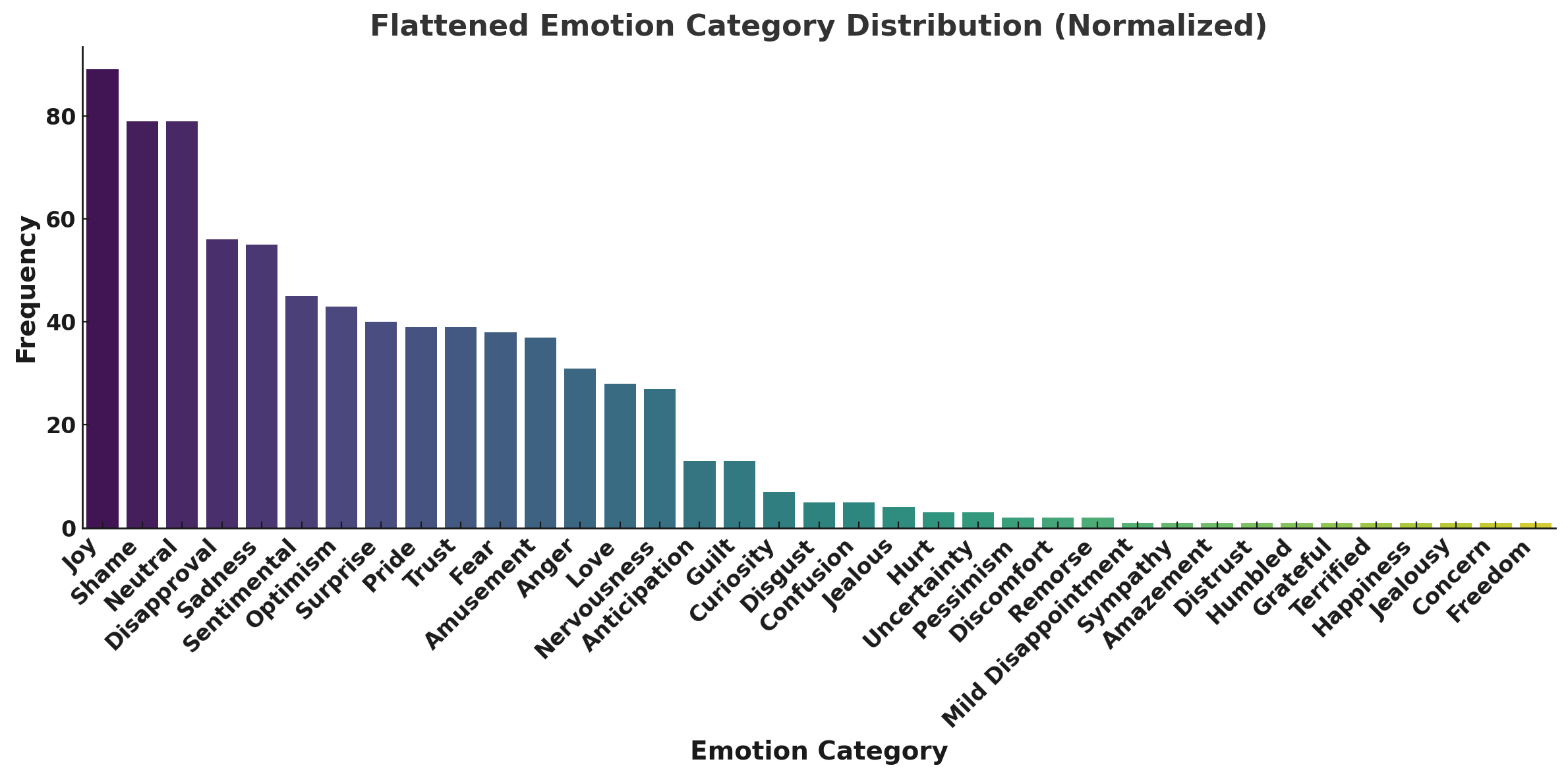}
\caption{Flattened Emotion Category Distribution (Normalized)}
\end{figure*}

\subsection{Emotion-Personality Correlation}
We computed Pearson correlations between normalized emotion categories and personality traits using both MBTI and OCEAN frameworks. Emotion labels were one-hot encoded, and personality traits were encoded either as binary (MBTI) or ordinal (OCEAN).

\textbf{MBTI Correlation:} Each of the 16 standard MBTI types (e.g., INFP, ESTJ) was represented as a binary feature. We then calculated the Pearson correlation between these types and each individual emotion category. This analysis highlights patterns such as higher emotional resonance of Joy with ENFP and stronger ties between Apprehension and ISFP types.

\begin{figure*}[t]
\centering
\includegraphics[width=0.9\textwidth]{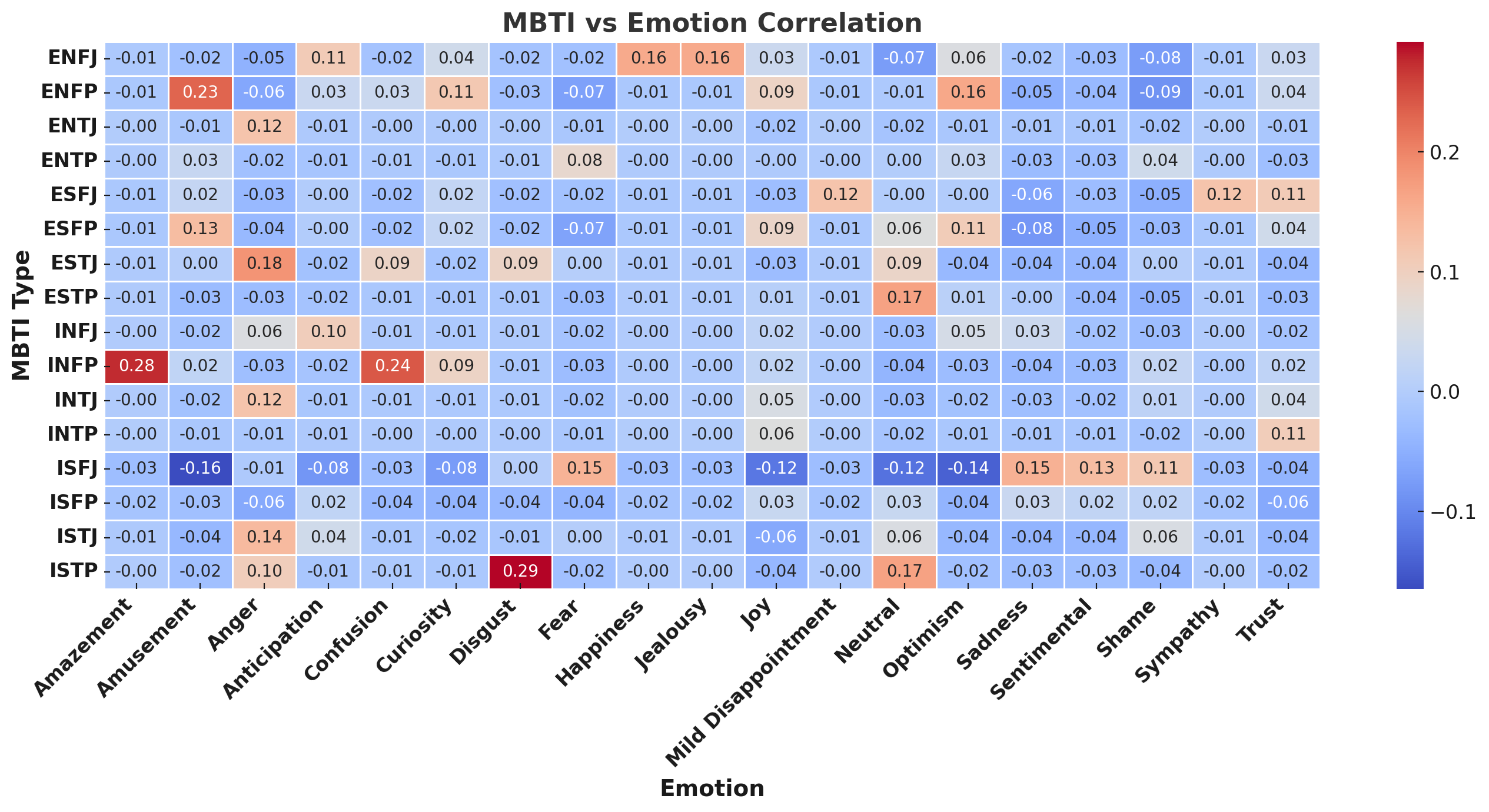}
\caption{MBTI vs Emotion Correlation}
\end{figure*}

\textbf{OCEAN Correlation:} OCEAN traits were originally qualitative and mapped numerically: Low = 1, Medium = 2, High = 3. We computed correlations between each trait and each emotion category. For instance, Neuroticism shows a positive correlation with emotions like Fear and Nervousness, while Agreeableness aligns more strongly with Trust and Caring emotions.

\begin{figure*}[t]
\centering
\includegraphics[width=0.9\textwidth]{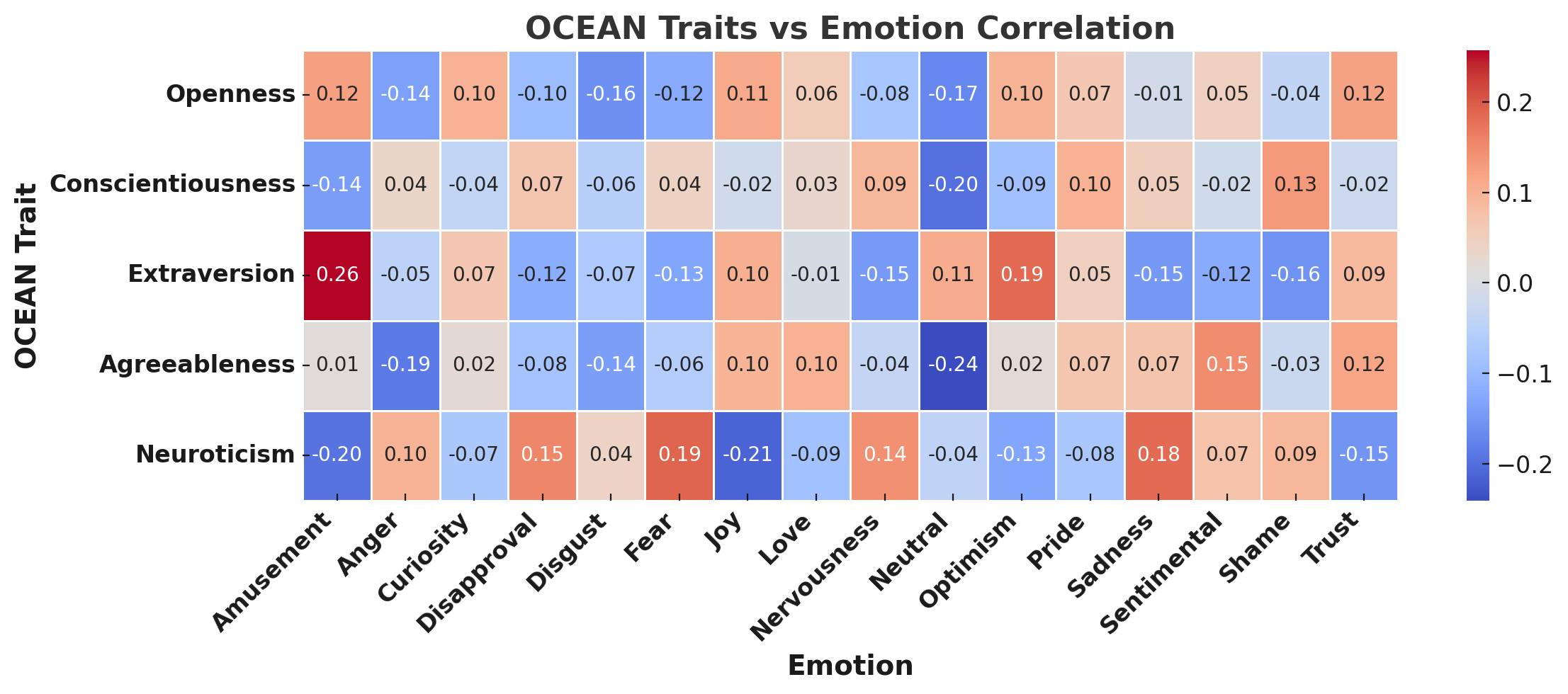}
\caption{OCEAN Traits vs Emotion Correlation}
\end{figure*}

\section{EU Task Categories}
\label{sec:task_definition}
We evaluate our framework across four emotional understanding (EU) task categories defined in EmoBench \cite{sabour-etal-2024-emobench}:

\begin{itemize}
    \item \textbf{Complex Emotions (CB)}: This task category evaluates a model’s ability to reason about nuanced and layered emotional experiences. It includes scenarios involving (1) emotional transitions in response to evolving events, (2) mixtures of emotions with potentially conflicting valence (e.g., joy and disappointment), and (3) unexpected emotional outcomes that challenge commonsense assumptions. These situations require deeper inferential reasoning to correctly interpret how emotions manifest across dynamic or atypical contexts.
    
    \item \textbf{Personal Beliefs and Experiences (PBE)}: This category evaluates how well a model understands that emotional responses are shaped by a person's cultural values, sentimental attachments, and prior experiences. Scenarios may involve culturally influenced norms (e.g., attitudes toward punctuality), differing levels of sentimental value assigned to objects or events, or reactions driven by personal traits and experiences, such as phobias or personality styles. Accurate prediction in this category requires sensitivity to how internal beliefs and past experiences modulate emotional appraisal.
    
    \item \textbf{Perspective-Taking (PT)}: This category focuses on the model’s ability to simulate others’ emotional states by reasoning about their beliefs, knowledge, and social context. It adapts classical theory-of-mind tasks: False Belief, Faux Pas, and Strange Stories to scenarios where emotional inference depends on understanding what different individuals know or assume. For instance, detecting excitement in someone misled by false information, or recognizing that embarrassment does not occur when a faux pas is unknown to the speaker, requires the model to attribute distinct perspectives to each character in the scenario.
    
    \item \textbf{Emotional Cues (EC)}: This task assesses a model’s ability to infer emotions from implicit textual descriptions of vocal and visual cues, such as tone, speech patterns, facial expressions, or bodily gestures. Unlike explicit emotional statements, these scenarios require the model to recognize subtle indicators. For instance, interpreting a sigh as annoyance or relief, or a flushed face as either anger or embarrassment, based on context. This category tests whether LLMs can perceive and reason about nonverbal affective signals conveyed through text.
\end{itemize}

\section{Prompt Templates}

This section outlines the prompt configurations used in our PTEI framework. Prompts are grouped by their corresponding modules.

\subsection{Prompts for Scenario Generation} \label{sec:prompt_scenario}

 The prompt template for scenario generation was developed independently of any benchmark items, as shown in Table~\ref{tab:scenario_prompt}.
\begin{table*}[t]
\centering
\caption{Scenario generation prompt template with structured goals, categories, and output fields.}
\small
\begin{tabular}{p{0.95\linewidth}}
\toprule
\textbf{Scenario Generation Prompt Template} \\
\midrule
You are a scenario generator for emotion understanding. \\[0.5em]

\textbf{Instruction:} \\
Generate a scenario designed to evaluate emotional reasoning and perspective taking. \\[0.5em]

\textbf{Goals and Categories:} \\
1. Perspective Taking: False Belief, Faux Pas, Strange Story \\
2. Emotional Cues: Vocal Cues, Visual Cues \\
3. Complex Emotions: Emotion Transition, Mixture of Emotions, Unexpected Outcome \\
4. Personal Beliefs and Experiences: Cultural Values, Sentimental Value, Persona \\[0.5em]

\textbf{Main Prompt Body:} \\
This situation should contain: \\
-- "Goal" (selected from above) \\
-- "Category" (with category description) \\
-- "Scenario" (short, coherent situation) \\
-- "Subject" (the person the scenario is about) \\
-- 5--6 "Choices" for Emotion (single or combined, from predefined list) \\
-- "Label" for correct Emotion \\
-- 4--6 "Choices" for Cause (may combine elements with ``\&'') \\
-- "Label" for correct Cause \\[0.5em]

\textbf{Constraints:} \\
Emotions must be selected from the predefined list of 50+ emotions. \\[0.5em]

\textbf{Expected Output (in JSON-like format):} \\
\texttt{\{ } \\
\texttt{\ \ \ "Goal": "...",} \\
\texttt{\ \ \ "Category": "...",} \\
\texttt{\ \ \ "Scenario": "...",} \\
\texttt{\ \ \ "Subject": "...",} \\
\texttt{\ \ \ "Emotion\_Choices": ["...", "..."],} \\
\texttt{\ \ \ "Emotion\_Label": "...",} \\
\texttt{\ \ \ "Cause\_Choices": ["...", "..."],} \\
\texttt{\ \ \ "Cause\_Label": "..."} \\
\texttt{\}} \\[0.5em]

\textbf{Task:} \\
Now generate a new example in the same format. \\
\bottomrule
\end{tabular}
\label{tab:scenario_prompt}
\end{table*}

\subsection{Prompts for Personality Detection Module} \label{sec:prompt_personality}
These prompts are used to infer personality traits for each subject within the scenario. 

\begin{itemize}
    \item MBTI Personality Prompt: Table~\ref{tab:mbti_prompt_2shot}
    \item OCEAN Personality Prompt: Table~\ref{tab:ocean_prompt_2shot}
\end{itemize}

\subsection{Prompts for EU Task}
These prompts are designed to guide LLMs in predicting emotions and their causes for each scenario, with or without step-by-step reasoning. They incorporate contextual retrieval and personality traits.

\begin{itemize}
    \item PTEI-Base Prompt: Table~\ref{tab:ptei_prompt}
    \item PTEI-CoT Prompt: Table~\ref{tab:ptei_cot_prompt}
\end{itemize}

\begin{table*}[t]
\centering
\caption{MBTI prompt template with structured instructions and two-shot demonstration format.}
\small
\begin{tabular}{p{0.95\linewidth}}
\toprule
\textbf{MBTI Personality Prompt Template (with 2-Shot Demonstrations)} \\
\midrule
You are a personality assessment expert trained to analyze behavioral patterns and assign MBTI personality types. \\[0.5em]
\textbf{MBTI Dimensions:} \\ 
-- I vs. E: alone/quiet vs. social/active \\
-- S vs. N: practical/details vs. abstract/ideas \\
-- T vs. F: logic/objectivity vs. emotions/values \\
-- J vs. P: structured/planned vs. flexible/spontaneous \\[0.5em]
\textbf{Scenario Context:} \\
Category: \{\texttt{category\_name}\} \\
Explanation: \{\texttt{category\_explanation}\} \\[0.5em]
\textbf{Task:} Analyze the scenario, infer the subject's MBTI type, and explain the reasoning. \\[0.5em]
\textbf{Input:} \\
Scenario: \{\texttt{scenario}\} \\
Subject: \{\texttt{subject}\} \\[0.5em]
\textbf{Expected Output (in JSON format):} \\
\texttt{\{ }\\
\texttt{\ \ \ "MBTI": "XXXX",}\\
\texttt{\ \ \ "Explanation": "Reasoning based on traits observed in the scenario."}\\
\texttt{\}} \\[0.5em]
\textbf{Few-shot Examples:} \\
Two illustrative demonstrations (scenario + subject + MBTI + explanation) are included before the test case to guide prediction. \\
\bottomrule
\end{tabular}

\label{tab:mbti_prompt_2shot}
\end{table*}

\begin{table*}[t]
\centering
\caption{OCEAN prompt template used for personality trait inference, including structured trait outputs, reasoning explanation, and a two-shot demonstration format.}
\small
\begin{tabular}{p{0.95\linewidth}}
\toprule
\textbf{OCEAN Personality Prompt Template (with 2-Shot Demonstrations)} \\
\midrule
You are a personality assessment expert trained to analyze behavioral patterns and assign OCEAN personality traits. \\[0.5em]
\textbf{OCEAN Dimensions:} \\
-- Openness: curiosity, creativity, interest in new experiences \\
-- Conscientiousness: organization, responsibility, reliability \\
-- Extraversion: sociability, talkativeness, assertiveness \\
-- Agreeableness: kindness, trust, cooperation \\
-- Neuroticism: emotional instability, anxiety, moodiness \\[0.5em]
\textbf{Scenario Context:} \\
Category: \{\texttt{category\_name}\} \\
Explanation: \{\texttt{category\_explanation}\} \\[0.5em]
This context helps interpret how the subject expresses themselves and reacts emotionally. \\[0.5em]
\textbf{Task:} Read the scenario and predict the subject's OCEAN traits. You should generate "Low", "Medium" or "High" for each trait. Then explain the reasoning based on the observed behaviors. \\[0.5em]
\textbf{Input:} \\
Scenario: \{\texttt{scenario}\} \\
Subject: \{\texttt{subject}\} \\[0.5em]
\textbf{Expected Output Format (JSON):} \\
\texttt{\{} \\
\hspace*{1em} \texttt{"Openness": "High",} \\
\hspace*{1em} \texttt{"Conscientiousness": "Medium",} \\
\hspace*{1em} \texttt{"Extraversion": "Low",} \\
\hspace*{1em} \texttt{"Agreeableness": "High",} \\
\hspace*{1em} \texttt{"Neuroticism": "Medium",} \\
\hspace*{1em} \texttt{"Explanation": "Brief justification of the above traits based on behavioral cues in the scenario."} \\
\texttt{\}} \\[0.5em]
\textbf{Few-shot Examples:} \\
Two example scenarios with annotated trait levels and explanations are included before the test case to guide prediction. \\
\bottomrule
\end{tabular}

\label{tab:ocean_prompt_2shot}
\end{table*}

\begin{table*}[t]
\centering
\caption{PTEI-Base prompt template for emotion and cause prediction, incorporating memory-based retrieval and structured personality conditioning (MBTI + OCEAN).}
\small
\begin{tabular}{p{0.95\linewidth}}
\toprule
\textbf{PTEI-Base Prompt Template (with Memory Retrieval and Personality Context)} \\
\midrule
\textbf{Instruction:} \\
\texttt{In this task, you are presented with a scenario, a question, and multiple choices. Please carefully analyze the scenario and take the perspective of the individual involved.} \\
\texttt{Provide only one single correct answer to the question and respond only with the corresponding letter. Do not provide explanations for your response.} \\[0.5em]

\textbf{Few-shot Memory Retrieval:} \\
If retrieval is enabled, the prompt includes a few top-k scenarios retrieved from the memory bank. Each includes: \\
-- Scenario description \\
-- Annotated emotion and cause labels \\
-- MBTI type \\
-- OCEAN trait levels (High, Medium, Low) \\[0.5em]

\textbf{Main Prompt Body (Emotion Task Example):} \\
\texttt{You are a personality and emotion analyst.} \\
\texttt{First, carefully read and understand the following similar situations. Pay attention to how the individuals reacted emotionally, their causes, and their personalities.} \\
\texttt{\{\{Retrieved Scenarios\}\}} \\
\texttt{---} \\
\texttt{After analyzing these similar cases, consider the new situation below carefully.} \\
\texttt{Scenario: \{\{scenario\}\}} \\
\texttt{Personality Information: \{\{personality string\}\}} \\
\texttt{Question: What emotion(s) would \{\{subject\}\} ultimately feel in this situation?} \\
\texttt{Choices: A. ..., B. ..., C. ..., ...} \\[0.5em]

\textbf{Personality Context:} \\
If available, the personality string is appended in the form: \\
\texttt{...considering the MBTI personality is ESTJ and the levels of OCEAN personalities are Openness: Medium, Conscientiousness: High, ...} \\[0.5em]

\textbf{Main Prompt Body (Cause Task Example):} \\
\texttt{Question: Why would \{\{subject\}\} feel \{\{emotion\}\} in this situation?} \\

\bottomrule
\end{tabular}
\label{tab:ptei_prompt}
\end{table*}

\begin{table*}[t]
\centering
\caption{PTEI-CoT prompt template for emotion and cause prediction, combining retrieval, personality traits, and step-by-step reasoning. The model is expected to output both a selected answer and an explanation.}
\small
\begin{tabular}{p{0.95\linewidth}}
\toprule
\textbf{PTEI-CoT Prompt Template (with Memory Retrieval, Personality, and Chain-of-Thought Reasoning)} \\
\midrule
\textbf{Instruction:} \\
\texttt{In this task, you are presented with a scenario, a question, and multiple choices. Please carefully pay close attention to the emotions and intentions. Analyze the scenario and take the perspective of the individual involved. Reason step by step by exploring each option's potential impact on the individual(s) in question} \\
\texttt{Think step-by-step to identify the correct answer. Provide both the selected answer (as a single letter) and a brief explanation justifying your choice.} \\[0.5em]

\textbf{Few-shot Memory Retrieval:} \\
If retrieval is enabled, the prompt includes top-k examples from the memory bank. Each includes: \\
-- Scenario description \\
-- Annotated emotion and cause labels \\
-- MBTI type \\
-- OCEAN trait levels (High, Medium, Low) \\[0.5em]

\textbf{Main Prompt Body (Emotion Task Example):} \\
\texttt{You are a personality and emotion analyst.} \\
\texttt{First, carefully read and understand the following similar situations. Pay attention to how the individuals reacted emotionally, their causes, and their personalities.} \\
\texttt{\{\{Retrieved Scenarios\}\}} \\
\texttt{---} \\
\texttt{After analyzing these similar cases, consider the new situation below carefully.} \\
\texttt{Scenario: \{\{scenario\}\}} \\
\texttt{Personality Information: \{\{personality string\}\}} \\
\texttt{Question: What emotion(s) would \{\{subject\}\} ultimately feel in this situation?} \\
\texttt{Choices: A. ..., B. ..., C. ..., ...} \\[0.5em]

\textbf{Personality Context:} \\
If provided, personality cues are appended: \\
\texttt{...considering the MBTI personality is INFP and the levels of OCEAN personalities are Openness: High, Conscientiousness: Medium, ...} \\[0.5em]

\textbf{Main Prompt Body (Cause Task Example):} \\
\texttt{Question: Why would \{\{subject\}\} feel \{\{emotion\}\} in this situation?} \\[0.5em]

\textbf{Expected Output Format:} \\
\texttt{Answer: B} \\
\texttt{Explanation: The subject is likely to feel this way because ... (based on emotional cues, personality traits, and scenario context).} \\

\bottomrule
\end{tabular}

\label{tab:ptei_cot_prompt}
\end{table*}


\end{document}